\documentclass[a4paper]{article}

\usepackage{INTERSPEECH2020}

\usepackage{times}
\usepackage{graphicx}
\usepackage[font = Large]{subcaption}
\usepackage{latexsym}
\usepackage{adjustbox}
\usepackage{amsfonts}
\usepackage{algorithm}
\usepackage{amsmath}
\usepackage{soul}
\usepackage{enumitem}
\usepackage{array}
\usepackage{multirow}
\usepackage{makecell}
\usepackage{booktabs}
\usepackage{boldline}
\usepackage{footnote}
\usepackage[title]{appendix}
\usepackage{color, colortbl}
\usepackage{url}

\title{An Effective Domain Adaptive Post-Training Method \\ for BERT in Response Selection}
\name{Taesun Whang$^{1*}\thanks{* Work performed while at Korea University}$ \,\, Dongyub Lee\textsuperscript{2} \,\, Chanhee Lee\textsuperscript{3} \,\, Kisu Yang\textsuperscript{3} \,\, Dongsuk Oh\textsuperscript{3} \,\, Heuiseok Lim\textsuperscript{3}}
\address{$^1$Wisenut Inc.\\$^2$Kakao Corp.\\$^3$Korea University}
\email{taesunwhang@wisenut.co.kr, jude.lee@kakaocorp.com,\\ \{chanhee0222,willow4,inow3555,limhseok\}@korea.ac.kr}

\begin{document}
\maketitle
\begin{abstract}
We focus on multi-turn response selection in a retrieval-based dialog system. In this paper, we utilize the powerful pre-trained language model Bi-directional Encoder Representations from Transformer (BERT) for a multi-turn dialog system and propose a highly effective post-training method on domain-specific corpus. Although BERT is easily adopted to various NLP tasks and outperforms previous baselines of each task, it still has limitations if a task corpus is too focused on a certain domain.  Post-training on domain-specific corpus ({\em{e.g.,}} Ubuntu Corpus) helps the model to train contextualized representations and words that do not appear in general corpus ({\em{e.g.,}} English Wikipedia). Experimental results show that our approach achieves new state-of-the-art on two response selection benchmarks ({\em{i.e.,}} Ubuntu Corpus V1, Advising Corpus) performance improvement by 5.9\% and 6\% on $R_{10}@1$.
\end{abstract}


\section{Introduction}
Human computer conversation system aims to have natural and consistent conversation. In general, dialog systems can be categorized into task-oriented dialog systems and non-task-oriented dialog systems ({\em{i.e.,}} chatbot systems). In this paper, we focus on one of non-task-oriented dialog systems, especially retrieval-based dialog systems, in which the task of predicting most likely next utterance from a set of candidates pool. While generative-based models provide a consistent but inaccurate response, retrieval-based models have a advantage of ensuring accurate answer.\\
\indent Previous studies of response selection utilized recurrent neural networks \cite{lowe2015ubuntu} or convolutional neural networks \cite{kadlec2015improved} to represent dialog context and response, and predict their relevance score based on their representations. While these models represent dialog as a sequential sentence, utterance-response matching models build utterance-level encodings and utilize attention mechanism to catch relevant words between two sentences \cite{wu2017sequential,zhang2018dua}. Although attention-based utterance-response matching models are highly effective for predicting relevant response, self-attention-based matching model based on the work of Vaswani et al. \cite{vaswani2017attention} strengthens encoding sophisticated segment representations \cite{zhou2018multi,tao2019one,yuan2019multi}. Also, a model based on contextualized language representations is also applied to response selection task in the work of Vig and Ramea \cite{vig2019comparison} and showed effectiveness of the pre-trained contextual language models. \\
\indent Recently, pre-trained contextualized language models, such as ELMo \cite{peters2018deep}, BERT \cite{devlin2018bert}, and XLNet \cite{yang2019xlnet}) are proposed and showed great improvements on a wide range of NLP tasks, such as natural language inference, named entity recognition, and question answering. Despite their huge success, they still have limitations to represent contextual information in dialog corpus, specifically domain-specific corpus, since it is trained on general corpora ({\em{e.g.,}} English Wikipedia and Book Corpus). For example, Ubuntu Corpus, which is the most used corpus for evaluating retrieval-based dialog system, contains a number of terminologies and ubuntu manuals that do not usually appear in general corpora ({\em{e.g.,}} apt-get, lsmod, and grep). Since the corpus is biased towards a certain domain, pre-trained contextualized language models are not able to fully represent dialog contexts. In addition, conversation corpus, such as Twitter and Reddits, is mainly composed of colloquial expressions and abbreviation which are usually grammatically incorrect. In response selection task, one approach of training a certain domain knowledge embeddings is proposed. Chaudhuri et al. \cite{chaudhuri2018improving} proposed a method building domain understandable embeddings by incorporating external knowledge ({\em{e.g.,}} ubuntu manual description). In other NLP task, such as aspect extraction and review reading  comprehension (RRC), there has been an attempt to learn domain specificity based on pre-trained model. Xu et al. \cite{xu2019bert} proposed BERT based post-training method for RRC to enhance domain-awareness. Since reviews and opinion-based texts have many differences compared to the original corpus of BERT, therefore, post-training of BERT with two powerful unsupervised objectives ({\em{i.e.,}} masked language model (MLM) and next sentence prediction (NSP)) on task-specific corpus enhance to produce task-awareness contextualized representations. Also, additional training on down-stream corpus shows effectiveness and performance improvements in various NLP tasks \cite{phang2018sentence,wang2019superglue}. \\
\indent In this work, we propose an effective post-training method for a multi-turn conversational system. To the best of our knowledge, it is the first attempt to adopt BERT$_{\textit{base}}$ on the most popular response selection benchmark, Ubuntu Corpus V1. We demonstrate that NSP is especially considered as an important task for response selection, since classifying whether given two sentences are {\em{IsNext}} or {\em{NotNext}} is the ultimate objective of response selection. Also, we append {\normalsize{\verb|[EOT]|}} to the end of utterance that model can learn relationships among utterances during the period of post-training. 
Furthermore, our approach outperforms previous state-of-the-art performance by 5.9\% on $R_{10}@1$. We also evaluate on the recently released data set in the dialog system technology challenges 7 (DSTC 7) \cite{dstc19task1} and outperforms the 1st place of the challenges.

\section{Related Work}
Lowe et al. \cite{lowe2015ubuntu} proposed a new response selection benchmark dataset, Ubuntu Corpus V1, with a dual encoder baseline model. They utilized a RNN based models, specifically a vanilla RNN, a long short-term memory (LSTM), and a bi-directional LSTM. The sequential representations of a dialog context and a response are encoded by the RNN-based encoder and are utilized to predict the probability score of a given response being the next utterance in a given dialogue context. Kadlec et al. \cite{kadlec2015improved} conducted experiments that apply convolutional neural networks to a dual encoder architecture that offers better performance than that of the vanilla RNN model. Even though each turn of the dialog should be considered in a multi-turn conversation, a given dialog context was regarded as a sequence in the previous baselines. To alleviate this, Zhou et al. \cite{zhou2016multi} proposed the MultiView model that encodes both word-level and utterance-level representations. However, it does not have the ability to fully capture the relevance of the dialog context and the response. \\
\indent Therefore, more recently, the utterance-level dialog-response matching models were proposed to enable the model to readily catch the relevance of the dialog and the response \cite{wu2017sequential,zhang2018dua}. Wu et al. \cite{wu2017sequential} proposed a sequential matching network, utilizing both word embeddings and their sequential representations encoded by gated recurrent unit (GRU) encoder to build matching matrices between the dialogue context and the response. Zhang et al. \cite{zhang2018dua} proposed a highly effective turns-aware aggregation methodology and exercised self-matching attention to fuse the representation of each utterance. Another approach to the dialogue-response matching model proposed by Dong and Huang (2018), is designed based on enhanced sequential inference model (ESIM) \cite{chen2017enhanced} model. ESIM achieved a good performance on a natural language inference task and is a proper model for discriminating if given two sentences are relevant or not. As an self-attention-based model (Vaswani et al., 2017) achieved significant performance improvements on various NLP tasks; it was also adapted for the response selection task \cite{zhou2018multi,tao2019one,yuan2019multi}. Zhou et al. \cite{zhou2018multi} used self-attention and cross-attention simultaneously so as to capture token-level dependencies as well as relevant segment pairs from a dialog and a response. One of most recent models, interaction-over-interaction model alleviates shallow interaction between dialog-response matching by stacking multiple interaction blocks \cite{tao2019multi}. Yuan et al. \cite{yuan2019multi} pointed out previous dialog-response matching models in terms of excessive use of dialog context information, therefore they proposed multi-hop selector to filter only relevant utterances to response. They achieved state-of-the-art results on three response selection benchmark datasets ({\em{i.e.,}} Ubuntu Corpus V1, Douban Corpus, and E-commerce).

\section{Our Approach}
\subsection{Domain Post-Training}
\label{ssec:bert-dpt}
BERT is designed to be easily applied to other nlp tasks with a fine-tuning manner. Since it is pre-trained on general corpus ({\em{e.g.,}} Wikipedia Corpora), it is insufficient to have enough supervision of task-specific words and phrases during the period of fine-tuning. To alleviate this issue, we post-train BERT on our task-specific corpora that helps model understand certain domain. The model is trained with two objectives, MLM and NSP, which are highly effective to learn contextual representations from the corpora. One example (Ubuntu Corpus) of domain post-training of BERT for response selection is described in Table 1.

\begin{figure}[t]\centering
\includegraphics[width=0.48\textwidth]{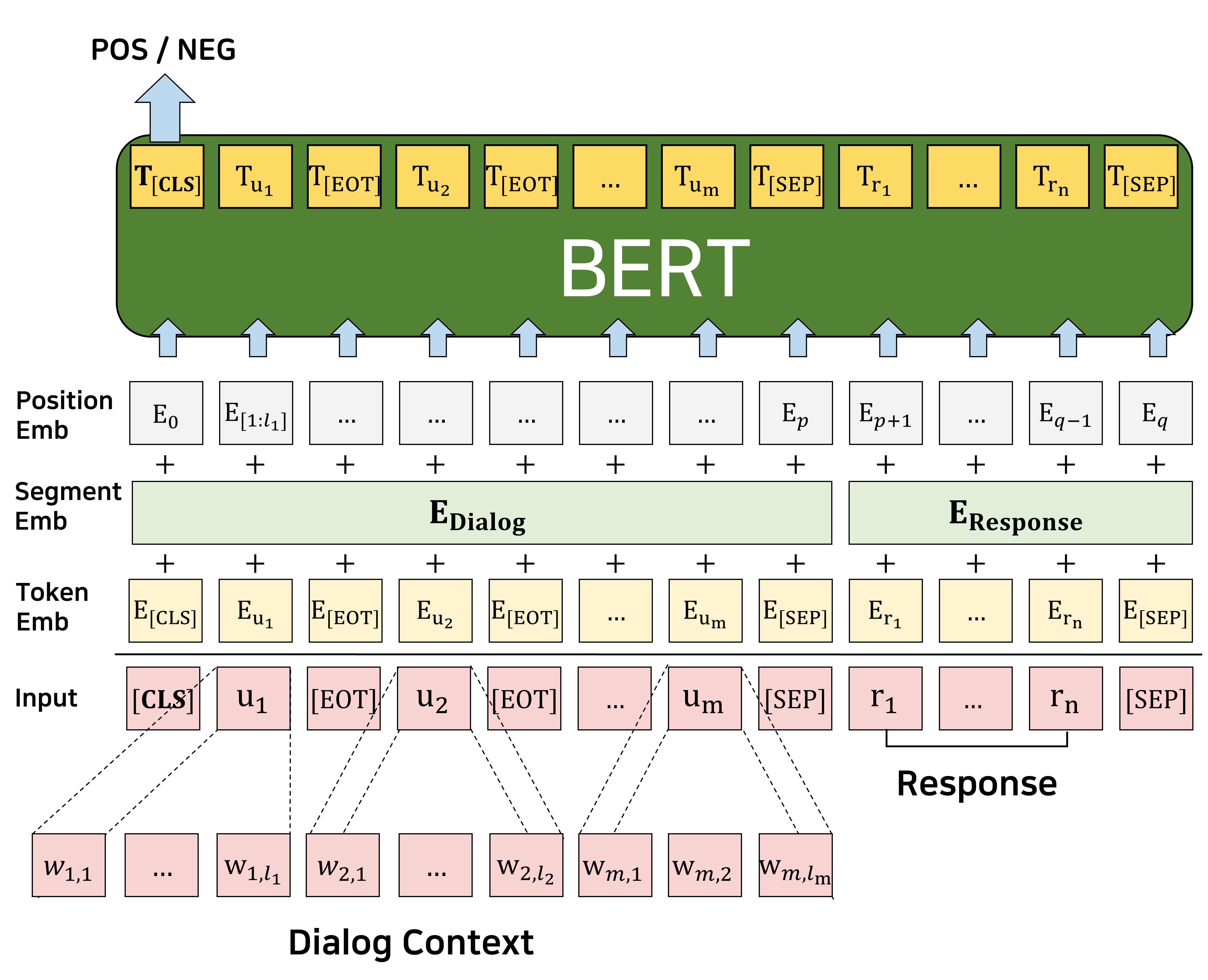}
\caption{BERT for Response Selection}
\label{fig:bert_response_selection}
\vspace{-0.3cm}
\end{figure}

In the example of Masked LM, model can learn that {\verb|sud|}\, {\verb|##o|} command is needed when trying apt install in Ubuntu system, which is not generally showed from universal corpora. Unlike general sentence, multi-turn dialog system is composed of a set of utterances. We append \textit{``\,End Of Turn\,''} token \normalsize{\verb|[EOT]|} to the end of each turn to make the model catch each utterance is finished at the point. By conducting NSP during the post-training, model also can train given two sentences are sequential and relevant, which is the common ultimate goal of response selection.
To optimize the model domain post-training (DPT) loss is calculated adding mean MLM likelihood and that of NSP, formulated as 
\begin{equation}
\mathcal{L}_{DPT} = \mathcal{L}_{MLM} + \mathcal{L}_{NSP}.
\label{eq:third_eq}
\end{equation}

\begin{table}[h]
\vspace{-0.2cm}
\begin{adjustbox}{width=0.4\textwidth}
\centering
\large
\begin{tabular}{l}
{\large\textbf{[Masked Language Model]}} \\
\textbf{Input :}

{\verb|sud|} {\verb|##o|} {\verb|apt|} {\verb|-|} {\verb|get|} {\verb|install|} {\verb|cuda|} \\\\

$\qquad\quad\,\Rightarrow$ {\verb|[MASK]|} {\verb|##o|} {\verb|apt|} {\verb|-|} {\verb|get|} {\verb|install|} {\verb|cuda|} \\\\

{\large\textbf{[Next Sentence Prediction]}} \\
\textbf{Input :}
{\verb|how|} {\verb|do|} {\verb|I|} {\verb|[MASK]|} {\verb|cuda|} {\verb|library|} \\$\qquad\quad\qquad\quad\quad${\verb|in|} {\verb|terminal|} {\verb|?|} {\verb|[EOT]|}{\verb|[SEP]|} (\textbf{Text A})\\\\
$\qquad\quad\,${\verb|You|} {\verb|can|} {\verb|search|} {\verb|with|} {\verb|apt|} {\verb|-|} {\verb|cache|} \\ $\qquad\quad\quad\quad\quad$ {\verb|search|} {\verb|[MASK]|} {\verb|[EOT]|} {\verb|[SEP]|} \,(\textbf{Text B})\\\\

\textbf{Label :} {\verb|IsNext|}

\end{tabular}
\end{adjustbox}
\label{tab:dialog_example}
\caption{An example of domain post-training in Ubuntu Corpus}
\vspace{-0.5cm}

\end{table}

\subsection{Fine-tuning BERT for Response Selection}
Our overall approach is described in Figure \ref{fig:bert_response_selection}. We transform the task of ranking responses from the candidates pool into binary classification problem by approaching a pointwise method. We denote a training set as a triples $\mathcal{D}=\{(c_i,r_i,y_i)\}_{i=1}^{N}$, where ${c}=\{u_{1}, u_{2}, ..., u_{m}\}$ is a dialog context consists of a set of utterances. An utterance ${u_{i}}=\{w_{i,1}, w_{i,2}, ..., w_{i,\textit{l}_{i}}\}$ is composed of a set of word tokens $w_{i,k}$, where $1\leq$ $k$ $\leq$ $l_{i}$ and $l_{i}$ is the length of $i$-th utterance. Response is denoted as $r_i=\{r_1, r_2, ..., r_n\}$ ($n$ is the number of tokens in the response), and ground truth ${y_i}\in\{0,1\}$. We define BERT input $\mathbf{x}$ $=$ ({\normalsize{\verb|[CLS]|}}, $u_{1}$, {\normalsize{\verb|[EOT]|}}, $u_{2}$, {\normalsize{\verb|[EOT]|}}, ..., $u_{m}$, {\normalsize{\verb|[SEP]|}},  $r_{1}$, ..., $r_{n}$, {\normalsize{\verb|[SEP]|})}. Maximum sequence length of the dialog context and response is denoted as $q$. Position, segment, and token embeddings are added and fed into the BERT layers. The BERT contextual representations of {\normalsize{\verb|[CLS]|}} token, $\mathbf{T}_{{\small{\verb|[CLS]|}}}$, is utilized to classify whether a given dialog context and response is {\normalsize{\verb|IsNextUtterance|}} or not. We feed  $\mathbf{T}_{{\small{\verb|[CLS]|}}}$ to single-layer perceptron to compute the model prediction score,
\begin{equation}
g(c,r) = \sigma(\mathbf{W}_{task}^{\top}{\mathbf{T}_{{\small{\verb|[CLS]|}}}}+b),
\label{eq:first_eq}
\end{equation}
\noindent where $\mathbf{W}_{task}$ is a task-specific trainable parameter. We use cross entropy loss as the objective function to optimize our model, formulated as
\vspace{-0.15cm}
\begin{multline}
\mathcal{L}oss = -\sum_{(c,r,y)\in\mathcal{D}}  ylog(g(c,r))\\ + (1-y)log(1-g(c,r)).\;\;\;
\label{eq:second_eq}
\end{multline}

\section{Experiments}
\subsection{Datasets and Training Setup}
We evaluate our model on two multi-turn dyadic data sets, Ubuntu IRC (Internet Relay Chat) Corpus V1\cite{lowe2015ubuntu} and Advising Corpus\footnote{\begin{footnotesize}\url{https://github.com/IBM/dstc-noesis}\end{footnotesize}} \cite{dstc19task1}. 
For the Ubuntu Corpus, training set is composed of 0.5M dialog context containing positive and negative response with the ratio of $1$:$1$. Each validation and test set contains 50k dialog context with one positive response and nine negative responses.
Advising Corpus consists of 100k dialogs for training set and 500 for validation and test set. All sets contain one positive response and 99 negative responses. We only use one negative sample for training to make same conditions with Ubuntu Corpus.
For an evaluation metric, we use $R_{n}@k$, evaluating if the ground truth exists in top k from $n$ candidates \cite{lowe2015ubuntu, wu2017sequential}. We also use another evaluation metric mean reciprocal rank (MRR) \cite{voorhees1999trec}. 

\subsection{Implementation Details}
The models are implemented using the Tensorflow library \cite{abadi2016tensorflow}.
We use the uncased BERT$_{base}$ model\footnote{\begin{footnotesize}\url{https://github.com/google-research/bert}\end{footnotesize}} as a base code for our experiments.
The batch size is set to 32 and the maximum sequence length is set to 320, specifically 280 for a dialog context and 40 for a response. We post-train the model more on Ubuntu Corpus V1 and Advising Corpus, 200,000 steps and 100,000 steps, respectively. The model is optimized using Adam weight decay optimizer with learning rate of 3e-5.

\subsection{BERT Models}
\textbf{BERT$_{\textit{base}}$} \cite{devlin2018bert} is a vanilla BERT$_{\textit{base}}$ model for response selection task. Uncased BERT$_{\textit{base}}$ model checkpoint is utilized as initial weights of the model, and single-layer perceptron are trained during the fine-tuning phase. \\
\textbf{BERT-DPT} is a BERT$_{\textit{base}}$ model that is post-trained on domain-specific corpus, Ubuntu Corpus V1 and Advising Corpus. The unsupervised objectives of the BERT model, MLM and NSP tasks are conducted during the DPT phase. \\
\textbf{BERT-VFT} is a model that selects the number of top layers to fine-tuning in BERT-DPT. We evaluate the model with varying $T=\{0, 2, 4, 6, 8, 10, 12\}$, where $T$ is the number of layers which are tuned during training time. When $T=12$, all the BERT-DPT layers are fine-tuned and only embedding layers are fine-tuned when $T=0$. We fill in the best score based on MRR in Table \ref{table:variable_fine_tuning}. \\
\textbf{BERT-VFT(DA)} performs data augmentation technique by increasing the number of negative samples. The number of negative samples are decided by several experiments, when the number is 4 shows the best results. Negative responses are randomly resampled for every epoch from response candidates pool.  

\begin{table}[h]\centering
\begin{adjustbox}{width=0.45\textwidth}
\centering
\renewcommand{\arraystretch}{1.2}%
\begin{scriptsize}
\begin{tabular}{cccc}
\hline

\multicolumn{1}{l|}{\textbf{Model}}   & $R_{10}@1$    & $R_{10}@2$   & $R_{10}@5$   \\
\hline
\multicolumn{1}{l|}{DualEncoder$_{\textit{rnn}}$}     & 0.403    & 0.547  & 0.819 \\
\multicolumn{1}{l|}{DualEncoder$_{\textit{cnn}}$}     & 0.549  & 0.684  & 0.896 \\
\multicolumn{1}{l|}{DualEncoder$_{\textit{lstm}}$}    & 0.638    & 0.784  & 0.949 \\
\multicolumn{1}{l|}{DualEncoder$_{\textit{bilstm}}$}     & 0.630  & 0.780  & 0.944 \\
\hline

\multicolumn{1}{l|}{MultiView}     & 0.662    & 0.801  & 0.951 \\
\multicolumn{1}{l|}{SMN}     & 0.726  & 0.847    & 0.961  \\
\multicolumn{1}{l|}{AK-DE-biGRU}    & 0.747  & 0.868  & 0.972  \\
\multicolumn{1}{l|}{DUA}     & 0.752  & 0.868    & 0.962  \\ 
\multicolumn{1}{l|}{DAM}     & 0.767  & 0.874    & 0.969  \\ 
\multicolumn{1}{l|}{MRFN}     & 0.786  & 0.886   & 0.976\\ 
\multicolumn{1}{l|}{IoI}     & 0.796  & 0.894    & 0.974  \\ 
\multicolumn{1}{l|}{MSN}     & \underline{0.800}  & \underline{0.899}    & \underline{0.978}  \\

\hline
\hline
\multicolumn{1}{l|}{BERT$_{\textit{base}}$}   & 0.817  & 0.904    & 0.977  \\ 
\multicolumn{1}{l|}{BERT-DPT} & 0.851  & 0.924  & 0.984 \\
\multicolumn{1}{l|}{BERT-VFT} & \textbf{0.855}  & \textbf{0.928}  & \textbf{0.985}  \\
\multicolumn{1}{l|}{BERT-VFT(DA)} & \textbf{0.858}  & \textbf{0.931} & \textbf{0.985 }\\
\hline
\end{tabular}
\end{scriptsize}
\end{adjustbox}
\caption{Model comparison on Ubuntu Corpus V1.}
\label{table:evaluation_results_ubuntu}
\vspace{-0.7cm}
\end{table}

\subsection{Performance on Ubuntu Corpus V1}
For Ubuntu Corpus V1, we compared our model with the following baseline methods and the results are given in Table \ref{table:evaluation_results_ubuntu}.  \\
\textbf{Dual Encoder} is simple dialogue-response matching model based on RNN, CNN, LSTM, and BiLSTM \cite{kadlec2015improved}. Each dialog context and response is represented by each encoder and utilized to obtain relevance score of given two sentences. \\
\textbf{MultiView} model is based on both token-level and utterance-level representations to catch utterance-level information in a dialog. CNN encoder is firstly used to encode each utterance representation, and Gated recurrent unit (GRU) encoder is used to encode both token-level representations and utterance-level representations \cite{zhou2016multi}. \\
\textbf{SMN} \cite{wu2017sequential} proposed utterance-response matching methods, specifically attention matrices created by word embeddings and sequential representations are encoded by CNN and they are fed into GRU encoder to obtain a probability score.  \\
\textbf{AK-DE-biGRU} \cite{chaudhuri2018improving} proposed a method incorporating domain knowledge ({\em{i.e.,}} Ubuntu manual description). Domain knowledge embeddings are created based on manual description, BiGRU encoder's forward and backward last hidden states are concatenated to build the embeddings. Pre-trained word embeddings and created domain knowledge embeddings are added and fed into an attention module.\\
\textbf{DUA} \cite{zhang2018dua} is a method of modeling utterance aggregation. By conducting turns-aware aggregation and gated self attention, the model can give weights to more relevant utterance with response. \\
\textbf{DAM} \cite{zhou2018multi} is based on transformer \cite{vaswani2017attention} encoder and apply both self-attention and cross-attention to obtain matching scores. Each attention matching score is aggregated by 3D matching image.\\
\textbf{MRFN} \cite{tao2019multi} proposed multi representation fusion network and highlighted the effectiveness of fusing strategy. They proposed a methodology to fuse multiple representations of words, contexts, and attention. \\
\textbf{IoI} \cite{tao2019one} build interaction blocks to help model conduct deep interactions between utterances and response. All representations extracted from the blocks are aggregated to obtain a relevance score.\\
\textbf{MSN} \cite{yuan2019multi} is a previous state-of-the-art model, which proposed multi-hop selector network to control which dialog context information is reflected for matching response. \\
\indent In Ubuntu Corpus V1, it is observed that BERT-VFT achieves new state-of-the-art performance and it obtains performance improvement by 5.5\%, 2.9\%, 0.7\% in terms of $R_{10}@k$, where $k$=$\{1,2,5\}$ respectively, compared to the previous state-of-the-art method ({\em{i.e.,}} MSN). Focusing on $R_{10}@1$ metric, the performance of a vanilla BERT$_{\textit{base}}$ is 0.817 and comparing our main approach, which is BERT-DPT, shows better result (improvement by 3.4\%). \\
\indent In addition, we especially point out comparing BERT-VFT with other self-attention-based models, such as DAM, IoI, and MSN, since all models are built with similar model architecture. Especially, domain-specific optimized BERT-VPT model shows performance improvement by 8.8\% in terms of $R_{10}@1$ compared to the general transformer based model ({\em{i.e.,}} DAM). In the case of MSN, even though it uses word2vec \cite{mikolov2013distributed} as pre-trained word embeddings, it shows comparable results with BERT$_{\textit{base}}$ model. However, BERT-VFT that understands domain specific corpus shows 5.5\% higher results in terms of $R_{10}@1$, which has significant gap between two models.

\subsection{Performance on Advising Corpus}

As shown in Table \ref{table:evaluation_results_advising}, we compare our approach with two existing baselines on Advising Corpus, proposed by Vig and Remea  \cite{vig2019comparison} and Chen et al. \cite{chen2019sequential} in DSTC 7. The former baseline evaluate BERT$_{\textit{base}}$ model on the Advising Corpus, but there is substantial performance difference from what we obtain. We believe that different implementation frameworks and hyperparameters, such as learning rates, are the main reason why performance differences exist between our work and that of Vig and Remea \cite{vig2019comparison}. The first place of the challenge was the work conducted by Chen et al. \cite{chen2019sequential}, BERT-VFT outperforms 6\% in terms of $R_{10}@1$. DPT shows its effectiveness on Advising Corpus, leading 3.4\% higher results in terms of $R_{10}@1$ compared to the BERT baseline. 

\begin{table}[h]\centering
\begin{adjustbox}{width=0.45\textwidth}
\centering
\renewcommand{\arraystretch}{1.2}%
\begin{large}
\begin{tabular}{l|cccc}
\hline

\textbf{Model}   & $R_{100}@1$    & $R_{100}@10$   & $R_{100}@50$  & MRR  \\
\hline
Vig and Remma \cite{vig2019comparison}     & 0.186    & 0.580  & 0.942 & 0.312 \\
Chen et al. \cite{chen2019sequential}     & 0.214    & 0.630  & \textbf{0.948} & 0.339 \\
\hline
\hline
BERT$_{\textit{base}}$   & 0.236  & 0.656    & 0.946  & 0.359 \\ 
BERT-DPT  & 0.270  & \textbf{0.668}    & 0.942  & 0.395\\
BERT-VFT & \textbf{0.274}  & 0.654  & 0.932 & \textbf{0.400}\\
BERT-VFT(DA) & 0.274  & 0.664 & 0.942 & 0.399 \\

\hline

\end{tabular}
\end{large}
\end{adjustbox}
\caption{Evaluation results on the Advising Corpus.}
\label{table:evaluation_results_advising}
\vspace{-0.7cm}
\end{table}

\subsection{Variable Fine-tuning}
Houlsby et al. \cite{houlsby2019parameter} proposed efficient fine-tuning approach that only updates a few top layers of BERT during fine-tuning period. They suggested that small data sets may be sub-optimal, when fine-tuning the whole BERT layers. Inspired by this assumption, we conduct an experiment (Table \ref{table:variable_fine_tuning}) varying $T$ layers which is fine-tuned during the period of training ($T$=$\{0,2,4,6,8,10,12\}$). The models achieve the best performance when $T$=$4$ and $T$=$6$, for Ubuntu and Advising, respectively. We demonstrate that utilizing this application is effective on not only small sets but also domain-specific sets. Variable fine-tuning also helps for reducing computational resources ({\em{e.g.,}} GPU memory) as well as showing performance improvements.\\

\begin{table}[h]\centering
\vspace{-0.4cm}
\begin{adjustbox}{width=0.35\textwidth}
\centering
\renewcommand{\arraystretch}{1.1}%
			
\begin{tabular}{c|cc}

\hline
\textbf{Variable Fine-Tuning} & \multicolumn{1}{c|}{\textbf{Ubuntu}} & \multicolumn{1}{c}{\textbf{Advising}} \\
\hline

\multicolumn{1}{c|}{All layers}  & \multicolumn{1}{c|}{0.907}   & 0.395\\
\multicolumn{1}{c|}{Top 10 layers}  & \multicolumn{1}{c|}{0.908}  & 0.393 \\
\multicolumn{1}{c|}{Top 8 layers}   & \multicolumn{1}{c|}{0.909}  & 0.389 \\
\multicolumn{1}{c|}{Top 6 layers}   & \multicolumn{1}{c|}{0.909}   & \textbf{0.400} \\
\multicolumn{1}{c|}{Top 4 layers}   & \multicolumn{1}{c|}{\textbf{0.910}} & 0.392 \\
\multicolumn{1}{c|}{Top 2 layers}   & \multicolumn{1}{c|}{0.907} & 0.387 \\
\multicolumn{1}{c|}{Only Embedding Layers}  & \multicolumn{1}{c|}{0.904} & 0.386 \\
\hline

\end{tabular}

\end{adjustbox}
\caption{Variable fine-tuning in BERT model. BERT-VFT model is utilized for this experiment and the model is evaluated using MRR metric.}
\label{table:variable_fine_tuning}
\vspace{-0.7cm}
\end{table}

\subsection{Data Augmentation}
\indent We also experiment how effective data augmentation is for both data sets. Formerly, both training sets contain $1$:$1$ ratio of positive and negative responses since the model is trained with a pointwise manner. We change the ratio of the samples to $1$:$k$, where $k$ is the number of negative samples, by increasing $k$. The process of how we select the number of negative samples is heuristic, and the best performance is obtained at ratio of $1$:$4$.

\subsection{MLM and NSP in BERT}
Table \ref{table:MLM_NSP_EOT} shows the effectiveness of two unsupervised objectives of BERT, MLM and NSP in Ubuntu Corpus V1. Also, it shows that appending special token {\normalsize{\verb|[EOT]|}} at the end of each utterance is highly effective for response selection task. During the DPT phase, MLM helps the model learn contextual representations from the target domain corpus and it is trained  sequential consistency of the given two sentences by conducting NSP.\\
\indent In the case of NSP, it does not influence much in improvement of performance because what NSP do during the DPT phase is the same task as the model do during the fine-tuning phase. While conducting only NSP is not helpful for corpus understanding, MLM leads significant improvement in performance regardless of using the special token. 

\begin{table}[h]\centering
\begin{adjustbox}{width=0.45\textwidth}
\centering
\renewcommand{\arraystretch}{1.1}%
\begin{large}
\begin{tabular}{lccccc}
\hline

\begin{tabular}[c]{@{}c@{}}Post-Training \end{tabular} & \begin{tabular}[c]{@{}c@{}}Special \\ Token\end{tabular} & $R_{10}@1$    & $R_{10}@2$   & $R_{10}@5$  & MRR  \\
\hline
NSP &  \multirow{3}{*}{\begin{tabular}[c]{@{}c@{}}without \\ EOT\end{tabular}} & 0.816  & 0.903    & 0.977  & 0.884 \\ 
MLM & & 0.834  & 0.916    & 0.981  & 0.896\\
MLM + NSP & & \textbf{0.839}  & \textbf{0.920}  & \textbf{0.982} & \textbf{0.900}\\
\hline
NSP &  \multirow{3}{*}{\begin{tabular}[c]{@{}c@{}}with \\ EOT\end{tabular}}  & 0.819  & 0.906    & 0.978  & 0.886 \\ 
MLM &   & 0.838    & 0.918  & 0.982 & 0.899 \\
MLM + NSP & & \textbf{0.851}    & \textbf{0.924}  & \textbf{0.984} & \textbf{0.907} \\

\hline

\end{tabular}
\end{large}
\end{adjustbox}
\caption{Comparison of MLM and NSP on Ubuntu Corpus V1. Experiments are conducted depending on the use of [EOT].}
\label{table:MLM_NSP_EOT}
\vspace{-0.7cm}
\end{table}
\section{Conclusion}
In this paper, a highly effective post-training method for a multi-turn response selection is proposed and evaluated. Our approach achieved new state-of-the-art results for two response selection benchmark data sets, Ubuntu Corpus V1 and Advising Corpus. For future work, we will utilize external domain knowledge, such as ubuntu manual description in Ubuntu Corpus or curriculum information in Advising Corpus, for enhancing domain understandings.

\bibliographystyle{IEEEtran}
\bibliography{mybib}

\end{document}